\documentclass[runningheads]{llncs}
\usepackage{graphicx}
\usepackage{amsmath,graphicx}
\usepackage{graphicx}
\usepackage{url}            
\usepackage{booktabs}       
\usepackage{amsfonts}       
\usepackage{nicefrac}       
\usepackage{microtype}      
\usepackage{bm}
\usepackage[table]{xcolor}

\begin{document}
\title{Razor SNN: Efficient Spiking Neural Network with Temporal Embeddings}
%
%

\author{Yuan Zhang${}^{1}$ \quad Jian Cao${}^{1} $ \quad Ling Zhang${}^{1} $
\thanks{Correspondence to: Jian Cao. This work was supported by the National Key Research and Development Program of China (Grant No. 2018YFE0203801).}
\\
\quad Jue Chen${}^{1}$
\quad Wenyu Sun${}^{2}$
\quad Yuan Wang${}^{3}$}
\authorrunning{Yuan Zhang et al.}

\institute{$^1$School of Software and Microelectronics, Peking University \\
$^2$Alibaba Group \quad
$^3$School of Integrated Circuits, Peking University}
\maketitle              
\begin{abstract}
The event streams generated by dynamic vision sensors (DVS) are sparse and non-uniform in the spatial domain, while still dense and redundant in the temporal domain. Although spiking neural network (SNN), the event-driven neuromorphic model, has the potential to extract spatio-temporal features from the event streams, it is not effective and efficient. Based on the above, we propose an events sparsification spiking framework dubbed as Razor SNN, pruning pointless event frames progressively. Concretely, we extend the dynamic mechanism based on the global temporal embeddings, reconstruct the features, and emphasize the events effect adaptively at the training stage. During the inference stage, eliminate fruitless frames hierarchically according to a binary mask generated by the trained temporal embeddings. Comprehensive experiments demonstrate that our Razor SNN achieves competitive performance consistently on four events-based benchmarks: DVS 128 Gesture, N-Caltech 101, CIFAR10-DVS and SHD.

\end{abstract}
\section{Introduction}

Event-based neuromorphic computation utilizes sparse and asynchronous events captured by DVS to represent signals more efficiently. Unlike RGB cameras, DVS encodes the time, location, and polarity of the brightness changes for each pixel at high event rates \cite{name14}. Although the events are sparse in the spatial domain, the streams they composed are dense in the temporal domain. This characteristic makes event streams hardly process directly through deep neural networks (DNNs), which are based on dense computation. Fortunately, spiking neural networks (SNNs) have an event-triggered computation characteristic that matches well with processing events. However, it is desirable to accelerate the SNN models to be more suitable and efficient for real-time events tasks and further improve accuracy. 

The dynamic mechanism owns attention recipes, which selectively focus on the most informative components of the input, and can be interpreted as the sensitivity of output to the variant input. As for SNNs, inspired by \cite{name10} and \cite{name15}, we propose the \textbf{temporal embeddings} combined with the dynamic mechanism for SNNs, to explore the unstructured and data-dependent strategy. Heaps of prior works \cite{name14,name15,name16} are dedicated to the spatial-wise attention. Different from above works, the temporal embeddings emphasize on dense ticks of event streams.

\begin{figure}[t]
  \centering
   \includegraphics[width=0.85\linewidth, height=0.45\linewidth]{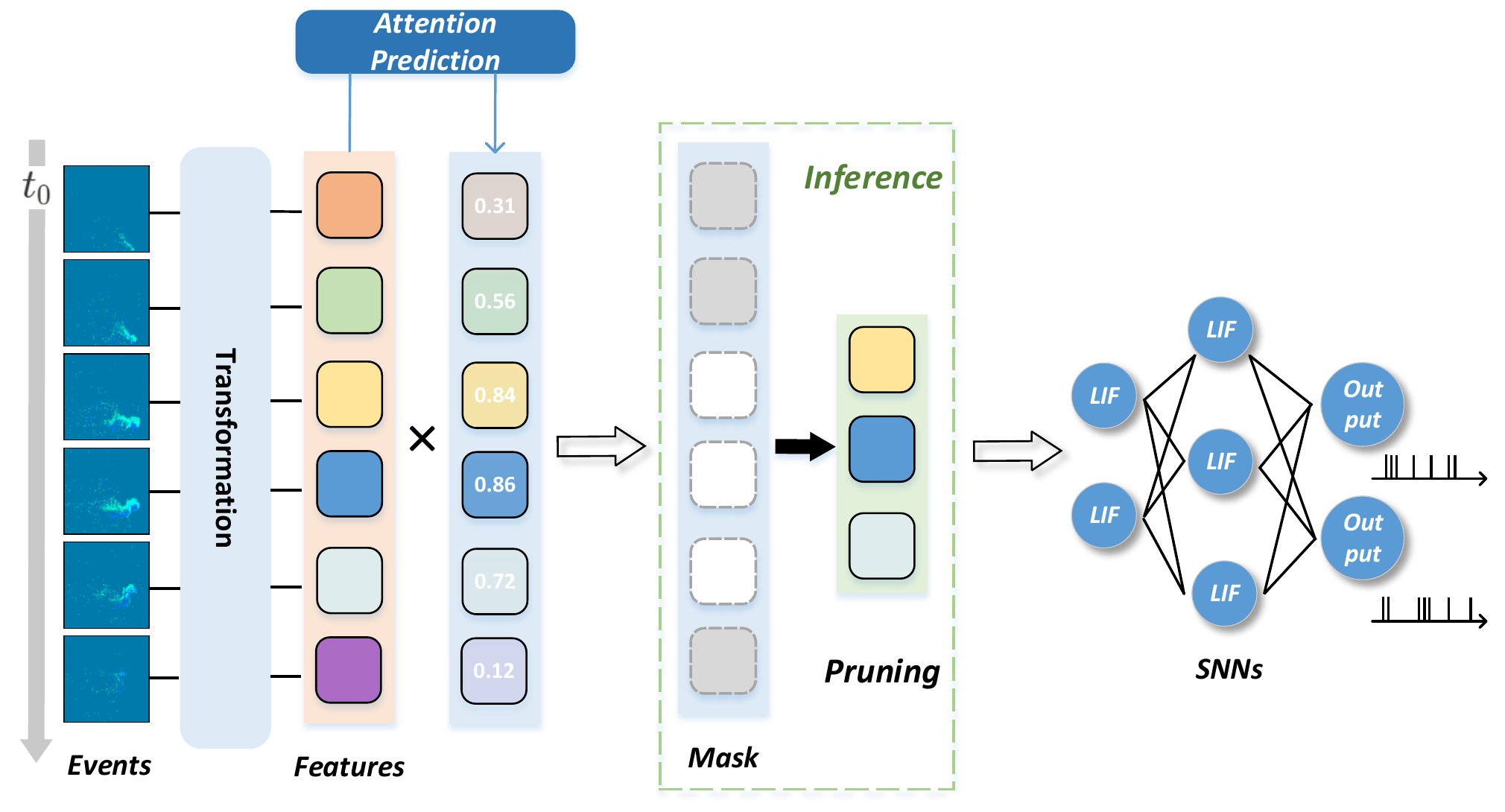}
   \caption{Event pruning for a spike layer in Razor SNN.}
   \label{fig:l}
\end{figure}

As shown in Fig.~\ref{fig:l}, we present an event pruning mechanism for the temporal domain by embeddings, to adaptively filter out immaterial events and get the SNNs slim. It would reconstruct features and predict attention vectors, to compute the probabilities of dropping the events, while retaining the event-triggered characteristic. We get the pruning architecture as Razor SNN. Our method reveals the possibility of exploiting the dynamic mechanism with temporal embeddings for the acceleration of SNNs and RNNs-like models. The contributions are listed as follows:
 
* Rethink the DVS events characteristics in both spatial and temporal aspects, and propose the novel pruning method named Razor SNN. It can do inference tasks with less data but higher performance. To the best of our knowledge, this is the first work to design a dynamic mechanism with temporal embeddings for SNNs.

* Our Razor SNN can achieve competitive performances on events recognition tasks, even compared with full-events inputs. Besides, it gets improvement of accuracy for gesture recognition with inference only 65 ms.

\section{Related Works}
\subsection{Object Recognition Using DVS}
For events recognition matched with a dynamic vision camera,  processing the events as groups is the most common method, to yield sufficient signal-to-noise ratios (SNR) \cite{amir2017low}. In this paper, we adopt the frame-based representation that accumulates events occured in a time window and maps into a frame \cite{name15}. It is convenient to generate frame-based representation and naturally compatible with the traditional computer vision framework.  Besides, SNN algorithms based on frames benefit from faster convergence for training \cite{perez2013mapping}. Timestep and window size is crucial to determine the quality of frame-based representation, the larger window size is, the higher SNR we could have. The prior works have been dedicated to taking various techniques to improve the classification performance based on the large window size. \cite{name7,name3} attempt to improve performance by taking training methods, while \cite{xu2018csnn,cheng2020lisnn} by changing the connection path of the SNNs, and \cite{deng2020rethinking,name6} through hybrid fusion.
\subsection{Spiking Neural Networks}
Based on the biologically plausible, the spiking neuron computes by transforming dynamic input into a series of spikes. Spike-based SNNs is to assume that the neurons which have not received any input spikes will skip computations, called event-triggered characteristic \cite{pfeiffer2018deep}.
There are heaps of research works on launching spiking neural networks for recognition tasks \cite{ruckauer2019closing}. Diehl et al. proposed a mechanism utilizing Spike Time Dependent Plasticity (STDP) \cite{caporale2008spike}, lateral inhibition and homeostasis to recognize. Lee et al. and Delbruck et al. have proposed a supervised learning mechanism mimicking back-propagation of conventional ANNs that can efficiently learn \cite{lee2016training}. The first one uses feed-forward layers of spiking neurons and a variant of back-propagation by defining an error function between desired and spiking activity. Wu et al. proposed Spatio-Temporal Backpropagation (STBP), a learning rule for back-propagating error in temporal and spatio domain \cite{name17}. This addresses the problem of approximating the derivative of the spike function that inherently brings in the question of biological plausibility. In this work, we adopt LIAF models as the elements of spike-based SNNs and STBP to evaluate the network architecture.
\subsection{Model Acceleration of SNNs}
There are various solutions that aim to compress or accelerate spiking neural networks. Structural network pruning methods explicitly prune out filters \cite{kundu2021towards}. Knowledge distillation methods \cite{zhang2021distilling,kushawaha2021distilling,zhang2023avatar} can guide the training of a student model with learnt knowledge, such as predictions and features, from a higher-capacity teacher (ANNs or SNNs). Some works design synchronous hardware inference mechanisms with parallelization strategies \cite{panchapakesan2021syncnn}. In our work, however, we aim to accelerate a SNN model based on feature map reconstruction with constraining width and depth of the model.

\section{Razor SNN}

\begin{figure*}[t]
  \centering
   \includegraphics[width=1.05\linewidth, height=0.41\linewidth]{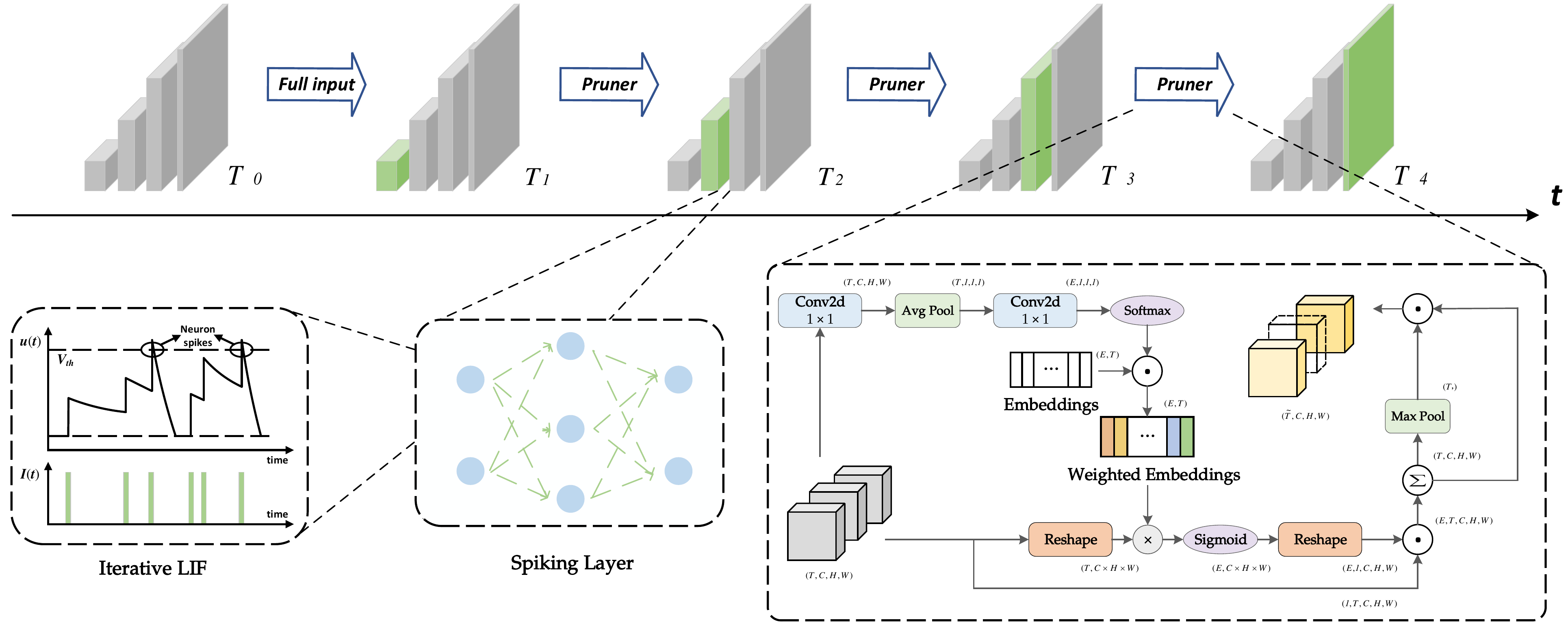}
   \caption{The complete forward flow of Razor SNN with Event Pruning Mechanism. The feature maps colored green are processed at timestamp $\bm{T}_i$. The dashed box in the bottom right corner represents the Event Pruning Mechanism we proposed. Zoom up to view better.}
   \label{fig:2}
\end{figure*}

\subsection{Iterative LIAF Model}

We first introduce the Leaky Integrate-and-Fire (LIF) model, a balance between complex dynamic characteristics and simpler mathematical form,  and translate it to an iterative expression with the Euler method \cite{name3}.  Mathematically it is updated as:

\begin{equation}
\bm{u}(t+1) = \tau \bm{u}(t) + \bm{I}(t),
  \label{eq:3}
\end{equation}

where ${\bm{u}\left(t\right)}$ denotes the neuronal membrane potential at time ${t}$, ${\tau}$ is a time constant for the leaky behavior and $\bm{I}\left(t\right)$  is the pre-synaptic input. The LIF model is as follows: 

\begin{equation}
\bm{a}_i^{t+1,\ l}=\sum_{j=1}^{l-1} \bm{W}_{i,j}^n \bm{O}_j^{t+1,l-1}.
  \label{eq:4}
\end{equation}

EQ.~\ref{eq:4} comes from the spatial domain. Where $\bm{a}_i$ is the axon inputs of the ${i}$th neuron, ${\bm{W}_{i,j}^n}$ is the synaptic weight from the ${j}$th neuron in previous layer to the ${i}$th neuron and $\bm{O}_j$ is the output of the ${j}$th neuron, whose value can only be 1 or 0. Besides, where the ${t}$ means the timestamp, ${l}$ represents the ${l}$th layer:
\begin{equation}
\bm{u}_i^{t+1,\ l}=\bm{u}_i^{t,l} \bm{h}\left(\bm{O}_i^{t,l}\right) + \bm{a}_i^{t+1,l}.
  \label{eq:5}
\end{equation}

EQ.~\ref{eq:5} comes from the temporal domain (TD). $\bm{u}_i$ is the membrane potential of the ${i}$th neuron. ${\bm{h}\left(x\right)}$ means the rate of TD memory loss as follows:

\begin{equation}
\bm{h}\left(x\right)=\tau e^{-\frac{x}{\tau}}.
  \label{eq:6}
\end{equation}

EQ.~\ref{eq:7} is the output of the ${i}$th neuron, responsible for checking whether the membrane potential exceeds the threshold $\bm{V}_{th}$ and firing a spike or not:

\begin{equation}
    \bm{O}_i^{t+1,\ l}=
    \left\{\begin{array}{l}
    {1} \;\; \qquad {u_i^{t+1,l} \geq V_{th}}, \\
    {0} \;\; \qquad {u_i^{t+1,l} < V_{th}}. \\
    \end{array}\right. 
    \label{eq:7}
\end{equation}

However, introducing the LIF model into the last layer will lose information on the membrane potential and disturb the performance. Instead, we adpot the leaky integrate-and-analog-fire (LIAF) model. LIAF changes the Heaviside step function to ReLU function, then both spatial and temporal domains are analog values.

\subsection{Event Pruning Mechanism}

The precise observation of significant events is the keystone to Dynamic Event Pruner. Unlike vanilla attention mechanisms, Razor SNN takes the information from neighboring frames into consideration with Global Temporal Embeddings. Besides, Prune the refined events to purify inputs.
For simplicity, we follow method \cite{name15} that the spatial input of ${l}$th layer at ${t}$th timestamp $\bm{O}^{t,\ l-1}$ equals to ${\bm{X}^{t,\ l-1} \in\mathbb{R}^{C\times H\times W}}$,  where $\bm{X}$ is the feature maps tensor and $C$ is channel size.

\subsubsection{Global Temporal Embeddings}
We introduce a set of learnable global temporal embeddings $\bm{B}\in\mathbb{R}^{E\times T}$ to extract $E$ sequence principles on temporal features.
Notably, not all embeddings play the same role. For example, we would assign more attention on the moment when events are dense, while less on that when events are sparse. In this paper, we propose an embedding weighting module to determine the embedding importance independently. 
Concretely, we conduct a convolution-based module with softmax function (see Figure 2) to predict, and weight the importance vector $\bm{w}$ onto the temporal embeddings to generate weighted embeddings $\hat{\bm{B}}$:

\begin{equation}
    \hat{\bm{B}}^{t} = \sum_{i=1}^T \bm{w_i} \odot \bm{B}^{t}.
\end{equation}

\subsubsection{Reconstruct Events Feature}
We accumulate the feature maps within $\bm{T}$ and flatten $\bm{X}$ into shape of $(T, C\times H\times W)$ in the paper. Then the Events of Interests (EoI) masks $\bm{M}$ can be obtained by calculating the similarities between weighted embeddings and the temporal frames in the feature maps: 

\begin{equation}
    \bm{M} = \sigma(\hat{\bm{B}}\bm{X}),
\end{equation}

where ${\sigma}$ denotes sigmoid activation function. Eventually, multiplying with the masks, we reconstruct Events Feature and get the refined feature $\hat{\bm{X}}$, which is a better substitute for the vanilla features:

\begin{equation}
    \hat{\bm{X}} = \sum_{i=1}^E \bm{M} \odot \bm{X}.
\end{equation}

\subsubsection{Pruning Strategy}

Due to the refined feature has contained discriminative temporal information provided by the Global Temporal Embeddings, it is what the pruning mechanism based on. We only need to send the refined feature to the adaptive max pooling 3d for extracting importance scores of temporal features $\bm{S} \in\mathbb{R}^{T}$, which is as follows:

\begin{equation}
\bm{S} = max_{i=1}^H max_{j=1}^W max_{k=1}^C \hat{\bm{X}},
\end{equation}

While during inference, we will generate a binary mask $\bm{M}$ according to the scores, where eliminate pointless frames which are lower than filtering threshold $\bm{S}_{th}$ we set, and set the attention score of the other frames to 1.

\begin{equation}
\bm{M}=\bm{H}(\bm{S}-\bm{S}_{th}).
\end{equation}

$\bm{H(\cdot)}$ is a Heaviside step function that is same as in  EQ.~\ref{eq:7}. Eventually, we combine the mask $\bm{M}$ with the input tensor, and the formed input at ${t}$th timestamp is:

\begin{equation}
\bm{\widetilde{X}}^{t,l-1}=\bm{M}^{t,l-1} \odot \hat{\bm{X}}^{t, l-1}.
\end{equation}

\subsection{Architecture Design of RazorSNNs}

We implement the RazorSNNs with embedding the Event Pruning Mechanism into each spiking layer except the encoder layer (the first layer). The reason is that, we assume SNN cannot extract more spatio-temporal information in this condition and pruning the whole network leads to unstable results. We follow the recent state-of-the-art methods\cite{wu2021training,name15} to use a x-stage pyramid structure, and the Razor Pruning Architecture is shown in Figure \ref{fig:2}.

\section{Experimental Results}

In this section, to show our method superiority and effectiveness, we conduct experiments on three popular event-based benchmarks: DVS Gesture, N-Caltech and CIFAR10-DVS.

\subsection{Implementation Details}
In this paper, we follow the similar notation as Yao et al. \cite{name15} to define our network architectures separately for DVS128 Gesture, SHD and CIFAR10-DVS, while N-Caltech's is the same as DVS128 Gesture's.
Besides, We take rate coding as loss function, and utilize the Adam optimizer for accelerating the training process. The corresponding hyperparameters details are shown in Tab \ref{tab:1}.

\begin{table}
  \caption{\textbf{Comprehensive parameters for experiments.}}
  \setlength\tabcolsep{3.0pt}
  \renewcommand\arraystretch{1.2}
  \label{tab:1}
  \centering
  \begin{tabular}{|l|l|l|}
    \hline
    Parameter & Description & Value \\
   \hline
    $\bm{dt}$ &	Window size	& ${1}$ms \\
    $\bm{V}_{th}$	& Fire threshold &	${0.3}$ \\
    $\bm{e}^{-\frac{dt}{\tau}}$ &	Leakage factor	& ${0.3}$ \\
    $\bm{S}_{th}$	& Razor ratio & ${0.4}$ \\
    \hline
  \end{tabular}
  \end{table}

\begin{table}
  \caption{\textbf{Accuracy of models for the SHD Dataset (\%). }}
  \label{tab:2}
  \setlength\tabcolsep{3.0pt}
  \renewcommand\arraystretch{1.3}
  \centering
  \begin{tabular}{c | c | c}
    \hline
   Methods	& Architecture & SHD \\
    \hline
        Cramer \cite{cramer2020heidelberg} & LIF RSNN & 71.40   \\ 
        Yin \cite{zenke2021remarkable} & RELU SRNN & 88.93  \\
        Zenke \cite{yin2020effective} & SG-based SNN & 84.00  \\
        Yao \cite{name15} & TA-SNN & 91.08 \\
        \rowcolor{blue!10}
        Ours & Razor SNN & \textbf{92.83}  \\
   \hline
  \end{tabular}
\end{table}

\subsection{Performance Comparison}
We compare Razor SNN with various of prior works for event-based data, like CNN method, spike-based SNN, and analog-based SNN, on above mentioned benchmarks. The experiment results are shown in Tab~\ref{tab:2}. From the results, our Razor SNN outperforms the strong baseline with a margin on SHD, CIFAR series and N-Caltech 101, and achieves competitive performance on DVS128 Gesture.

\textbf{SHD} \quad The SHD dataset \cite{cramer2020heidelberg} is a large spike-based audio classification task that contains 10420 audio samples of spoken digits ranging from zero to nine in English and German languages. Unlike the fourdimensional event stream generated by the DVS camera, the audio spike stream has only two dimensions, i.e., time and position. Our method surpasses the previous state-of-the-art by 1.75\%, and verify
the effectiveness of the Temporal embedding module.

\textbf{DVS128 Gesture} \quad Almost all the spike-based SNNs methods evaluate their model on it. Razor SNN surpasses TA-SNN by 0.22\% and outperforms all CNN-based methods. Due to unique temporal feature extraction and reconstruction, Razor SNN has superior ability to distinguish the clockwise and counterclockwise of gestures, which are easily confused.

\textbf{N-Caltech 101} \quad N-Caltech is a spiking version of frame-based Caltech101 dataset, which is a large-scale Events dataset. Few methods tested on N-Caltech because of the complicate background and computational complexity. Notably,  Razor SNN still gets a nearly 2.1\% increase over the TA-SNN, and outperforms all the SOTA methods. Besides, the temporal embeddings function as the attention module when SNNs meet static image, which is beneficial to classification.

\textbf{CIFAR10-DVS} \quad Compared to the best result TA-SNN so far, our method gets 1.01\% improvement, attributed to its global temporal module catching critical events information and filter temporal noise, which damages SNNs accuracy. Our Razor SNN shows better generalization on large-scale data set, where exists more noisy and pointless events.

Moreover, the number of parameters in Razor SNN only has weak increase compared with the vanilla SNN. So the events pruning mechanism can afford SNNs to achieve higher performance with less cost for practical applications.

\begin{table}
  \caption{\textbf{Comparison of different methods on DVS-Gesture, N-Caltech 101 and CIFAR10-DVS (\%). }}
  \label{tab:2}
  \setlength\tabcolsep{3.0pt}
  \renewcommand\arraystretch{1.3}
  \centering
  \begin{tabular}{c c | c c c}
    \hline
   Methods	& Architecture & Gesture & N-Caltech & CIFAR10\\
    \hline
        Wu \cite{name3} & NeuNorm & - & - & 60.50   \\ 
        Ramesh \cite{name4}  & DART & - & 66.8 & 65.78 \\
        Kugele \cite{name6} & DenseNet & 95.56 & - & 66.75 \\
        Wu \cite{name5} & LIAF-Net & 97.56 & - & 70.40  \\
        Zheng \cite{name7}& ResNet19 & 96.87 & - & 67.80  \\
        Kim \cite{kim2021optimizing} & SALT & 67.10 & 55.0 & - \\
        Wu \cite{wu2021training} & ASF-BP & 93.40 & 60.23 & 62.50  \\
        Yao \cite{name15} & TA-SNN & 98.61 & 68.42 & 72.00 \\
        \rowcolor{blue!10}
        Ours & Razor SNN & \textbf{98.83} & \textbf{70.50} & \textbf{73.01}  \\
   \hline
  \end{tabular}
  \end{table}

\subsection{Ablation Studies}


 \subsubsection{The Number of Temporal Embeddings} We perform experiments on DVS Gesture to explore the effects of different numbers of temporal embeddings in Razor SNN. As shown in Tab~\ref{tab:ab_num_tokens}, with only $2$ embeddings, Event Pruning improves SNN by $0.84\%$ Acc, while more tokens could achieve further improvements. In this paper, we choose a number of $\bm{4}$ for a better performance.
 
 \begin{table}[h]
	\renewcommand\arraystretch{1.8}
	\setlength\tabcolsep{5pt}
	\centering
	\caption{\textbf{Ablation on the number of temporal embeddings.}}
	\label{tab:ab_num_tokens}
	\footnotesize
	\begin{tabular}{ccccc}
	    \hline
	    0 (vanilla) & 2 & 4 & 6 & 8 \\
	    \hline
	    97.99 & 98.56 & $\textbf{98.83}$ & 98.79 & 98.34 \\
	    \hline
	\end{tabular}
\end{table}

\subsubsection{The Position of Where to Prune}
\begin{figure*}[ht]
  \centering
   \includegraphics[width=0.5\linewidth, height=0.375\linewidth]{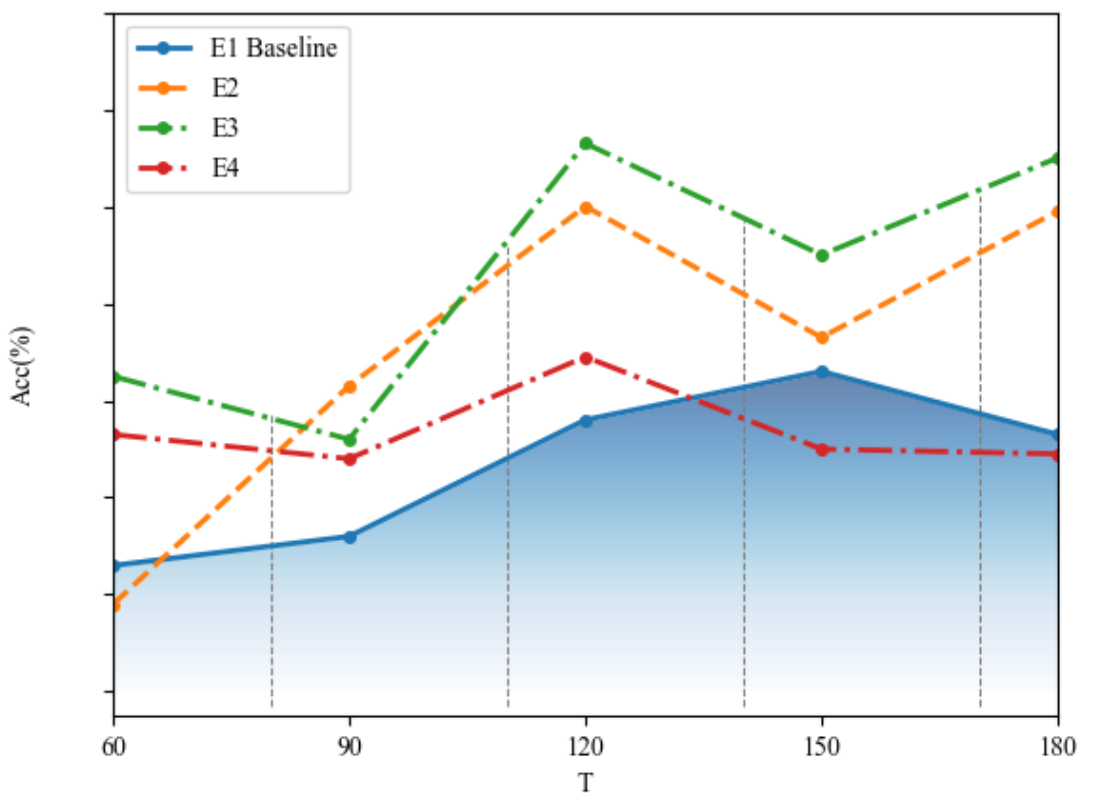}
   \caption{E1, pure SNNs baseline; E2, introduce embeddings into the encoder layer; E3, introduce embeddings into the backbone layers. E4, introduce embeddings into all the layers.}
   \label{fig:5}
\end{figure*}

It is vital to figure out that we should insert temporal embeddings into which layers to prune, and we design four sets of experiments to explore. E1, pure SNNs baseline; E2, introduce temporal embeddings into the encoder layer; E3, introduce temporal embeddings into the backbone layers. E4, introduce temporal embeddings into all the layers. As shown in Fig \ref{fig:5}, we observe that E2 and E3 independently afford improvement in most cases, where E3 achieves the best accuracy when T=120. But E4, who owns E2 and E3 simultaneously, leads to unstable results, and we assume SNNs cannot extract more spatio-temporal information in this condition.

\subsubsection{Effects of components in Event Pruning Mechanism} We set experiments to show the contribution of each proposed component in Mechanism in Tab~\ref{tab:ab_component}.
\textbf{+ Embeddings.} Global temporal embeddings benefit most ($0.31\%$) for Razor SNNs due to its consideration of neighboring frames and extraction of temporal features.
\textbf{+ Embeddings weighting module.}
Embedding weighting module decides the embedding importance independently and provide discriminative information and gains by $0.24\%$.
\textbf{+ Reconstruct Events Feature.}
The refined feature has contained discriminative temporal information, and experiment statics ($0.10\%$ ) proves that it a better substitute for the original features indeed.
\textbf{+ Pruning.} 
Pruning eliminates worthless events contained much noise which disturb the SNNs model.

\begin{table}
	\renewcommand\arraystretch{1.2}
    \setlength\tabcolsep{3pt} 
    \centering
    \caption{\textbf{Ablation experiments on effects of components in Event Pruning Mechanism.}}
    \label{tab:ab_component}
    \begin{tabular}{c c c c c}
    \hline
        Embeddings & Weighting & Reconstruct & Pruning & Acc(\%)\\
    \hline
        - & - & - & - & 97.99 (baseline) \\
        \checkmark & - & - & - & 98.30 (+$\bm{0.31})$ \\
        \checkmark & \checkmark & - & - & 98.54 (+0.24) \\
        \checkmark & \checkmark & \checkmark & - & 98.64 (+0.10)  \\
        \checkmark & \checkmark & \checkmark & \checkmark & 98.83 (+0.19)  \\
    \hline
  \end{tabular}
\end{table}

\subsection{Visualization Analysis}

\begin{figure}[hbt]
  \centering
   \includegraphics[width=0.4\linewidth, height=0.4 \linewidth]{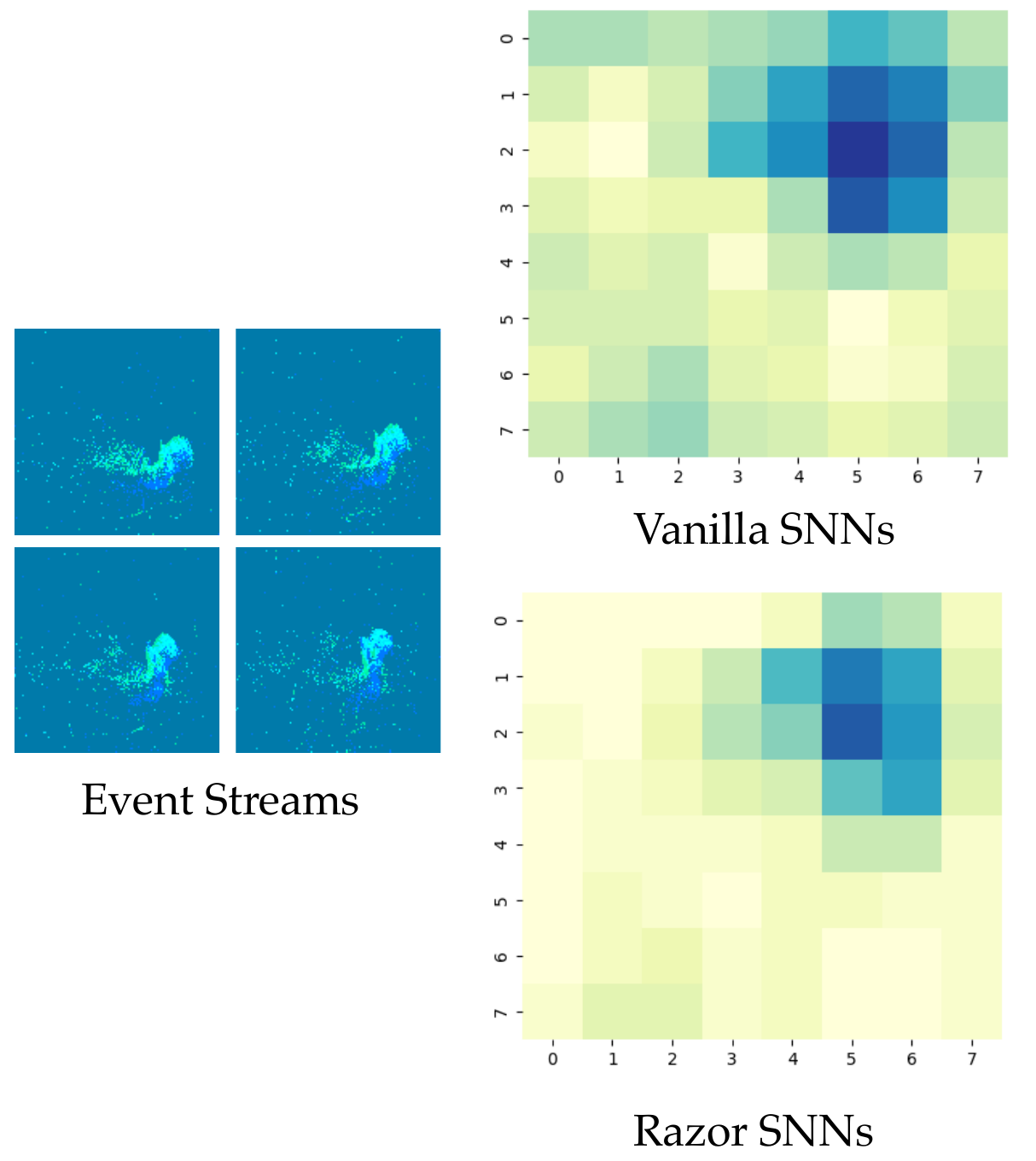}
   \caption{Visualization of the heat maps generated by vanilla SNNs and Razor SNNs separately. The temporal embeddings and reconstructed features urge SNNs to centre on the gesture instead of distraction somewhere like vanilla models. Best viewed in color.}
   \label{fig:vis}
\end{figure}

To validate the effectiveness of Razor SNN, we visualize the case that when the vanilla SNN fails in recognition, the Razor SNN succeeds. As shown in \ref{fig:vis}, each feature indicates the average response of a spiking layer. We make the following two observations about the effect of temporal embeddings and reconstruction of Razor SNN.

The spiking activity is more concentrated in Razor SNN, i.e., the deep blue area of Razor SNN is smaller and more focused. This suggests that global temporal extraction is beneficial to handling the important region of intermediate channels. We observe that the pruning lightens the color of the yellow area (background). The lighter the pixel, the weaker the spiking activity rate.

\section{Conclusion}
\label{sec:Conclusion}

In this paper, we innovatively introduce dynamic attention mechanism based on temporal embeddings into SNNs, and propose the Razor SNNs. Compared with vanilla Spiking Neural Networks, Razor SNNs process signals more efficiently, especially for event streams, pruning pointless event frames progressively. Our method enjoys finer temporal-level feature and prunes worthless events frames. Extensive experiments show that, our Razor SNNs apply to various benchmarks and can achieve state-of-the-art performance consistently.

%
%
%
\bibliographystyle{splncs04}
\bibliography{strings}
\end{document}